\documentclass[letterpaper]{article} 
\usepackage{aaai2026}  
\usepackage{times}  
\usepackage{helvet}  
\usepackage{courier}  
\usepackage[hyphens]{url}  
\usepackage{graphicx} 
\usepackage{natbib}  
\usepackage{caption} 
\usepackage{subcaption}
\usepackage{dblfloatfix}
\usepackage{booktabs}
\usepackage{algorithm}
\usepackage{algorithmic}
\usepackage{siunitx}
\usepackage{newfloat}
\usepackage{listings}
\DeclareCaptionStyle{ruled}{labelfont=normalfont,labelsep=colon,strut=off} 
\floatstyle{ruled}
\newfloat{listing}{tb}{lst}{}
\floatname{listing}{Listing}
\title{Reason-Plan-ReAct: A Reasoner-Planner Supervising a ReAct Executor for Complex Enterprise Tasks}
\author {
    Gianni Molinari\textsuperscript{\rm 1},
    Fabio Ciravegna\textsuperscript{\rm 1}
}
\affiliations {
    \textsuperscript{\rm 1}Università Degli Studi di Torino\\
    Via Pessinetto 12\\
    Turin, Italy\\
    gianni.molinari@unito.it, fabio.ciravegna@unito.it
}

\usepackage{bibentry}

\begin{document}

\maketitle

\begin{abstract}
Despite recent advances, autonomous agents often struggle to solve complex tasks in enterprise domains that require coordinating multiple tools and processing diverse data sources. This struggle is driven by two main limitations. First, single-agent architectures enforce a monolithic plan-execute loop, which directly causes trajectory instability. Second, the requirement to use local open-weight models for data privacy introduces smaller context windows leading to the rapid consumption of context from large tool outputs. To solve this problem we introduce RP-ReAct (Reasoner Planner-ReAct), a novel multi-agent approach that fundamentally decouples strategic planning from low-level execution to achieve superior reliability and efficiency. RP-ReAct consists of a Reasoner Planner Agent (RPA), responsible for planning each sub-step, continuously analysing the execution results using the strong reasoning capabilities of a Large Reasoning Model, and one or multiple Proxy-Execution Agent (PEA) that translates sub-steps into concrete tool interactions using a ReAct approach. Crucially, we incorporate a context-saving strategy within the PEA to mitigate context window overflow by managing large tool outputs via external storage and on-demand access. We evaluate RP-ReAct, on the challenging, multi-domain ToolQA benchmark using a diverse set of six open-weight reasoning models. Our empirical results show that RP-ReAct achieves superior performance and improved generalization ability over state-of-the-art baselines when addressing diverse complex tasks across the evaluated domains. Furthermore we establish the enhanced robustness and stability of our approach across different model scales, paving the way for effective and deployable agentic solutions for enterprises.  

\end{abstract}

 \begin{links}
     \link{Code}{https://github.com/giargiapower/RP-ReAct}
 \end{links}

\section{Introduction}
Autonomous agents are increasingly essential for executing complex tasks that require advanced reasoning and interaction with diverse tools~\cite{wu2024avatar, rivkin2024aiot, yao2023react}. Such capabilities are particularly critical in enterprise settings, where employees routinely interact with multiple systems and must reason over varied sources of information such as documents and databases \cite{muthusamy2023towards}.
LLM agents offer the ability to build adaptive, role-aware companions that enhance individual enterprise workers' productivity by assisting them in daily tasks.
Successfully achieving these goals requires robust, dynamic planning and a reliable execution mechanism for interacting with various functions.
Standard approaches are based on a single-agent which often struggle to maintain efficiency and trajectory in complex environments due to various challenges \cite{zhang2024chain, xu2023rewoo}.  First, the inherent difficulty of tool execution, which often leads to errors that must be corrected. Second, tool interactions that return vast amounts of data that can quickly fill the context window of the model.
Finally, the need for data privacy requires companies to deploy open-weight models internally. These models normally offer smaller context windows and lower generalization capability than proprietary counterparts.
So, when combined with the high data volume from error handling and tool execution, this results in a dual problem:
\begin{itemize}
    \item Context Consumption: This is one of the main problem when we have models with limited context window.
    \item Trajectory Deviation: The overload of information and internal error handling increases the risk of the agent deviating from the correct path toward the final goal.
\end{itemize}
To mitigate the computational cost and stability issues of monolithic agentic approaches, recent research has explored multi-agent architectures that separate planning and execution into specialized components \cite{zhang2025webpilot, erdogan2025plan}. This division not only reduces the initial context load but also minimizes the exposure to potential distractors during both critical phases. Furthermore, enabling Large Reasoning Models (LRMs) to incorporate an agentic search workflow into their decision-making has been shown to boost performance significantly in various cognitive tasks \cite{li2025search, uesato2022solving}.

Building upon these insights, this work introduces RP-React (Reasoner Planner-React), a novel multi-agent architecture designed to maximize efficiency and reliability in complex enterprise tasks. RP-React consists of:
\begin{itemize}
    \item A Reasoner Planner Agent (RPA): Responsible for receiving the complex task, leveraging powerful reasoning models to plan each sub-step toward the goal, and continuously analysing the execution results.
    \item Multiple Proxy-Execution Agents (PEA): Dedicated to translating the sub-step into concrete tool interactions, using a React approach (think, act, observe) to complete the action, and returning the result to the RPA.
\end{itemize}

This iterative cycle allows the RPA to continuously reason based on the PEA's responses, deciding whether to proceed to the next planned step or dynamically re-plan the trajectory in the event of an error or unexpected result. Crucially, we also introduce a context-saving strategy that directly addresses the context window overflow issue arising from large tabular tool outputs (SQL, CSV). when tool output exceeds a fixed threshold ($T$), a mechanism is defined for offloading externally part of the context of the PEA, so as to leave the context uncluttered. The offloaded part can however be uploaded or analysed on demand.
To test our agent's capabilities in an enterprise-like setting that requires multi-tool interaction and complex reasoning, we chose the ToolQA benchmark~\cite{zhuang2023toolqa}. We evaluated RP-ReAct's performance across five distinct domains, incorporating both easy and hard difficulty levels. Our results demonstrate that RP-React achieves high performance against state-of-the-art approaches, particularly in the most complex scenarios requiring intensive reasoning and numerous sequential tool calls. Critically, we demonstrate the architecture's robustness and stability across a variety of open-weight reasoning models.
Our key contributions are summarized as follows:
\begin{itemize}
\item We introduce RP-React, a multi-agent architecture that fundamentally separates the high-level strategic Reasoner Planner from low-level Proxy-Execution Agents, enabling dynamic re-planning and error handling.
\item We conduct a comprehensive evaluation across five ToolQA domains, including both easy and hard difficulty tasks, testing RP-React's approach on six diverse open-weight reasoning models.
\item We demonstrate that RP-React achieves superior performance, robustness, and generalization compared to state-of-the-art counterparts in hard tasks, offering a reliable path for integrating autonomous agents into complex enterprise automation tasks.
\end{itemize}

\section{Related Works}
\paragraph{\textbf{LLM Agents}}
Large Language Models (LLMs) are the core cognitive engine of modern agent architectures, driven by their robust natural language understanding and complex zero-shot reasoning capabilities~\cite{molinari2025towards}.
They are used to interact within complex environments~\cite{xu2024lac, zhao2024see} and tools~\cite{yao2023react, wu2024avatar, alakuijala2025memento, madaan2023self, rivkin2024aiot}.

Building upon the foundational capabilities of individual agents, various studies have explored the utility of cooperative multi-agent systems (MAS) to address tasks that can be too challenging for a 
single LLM. These collaborative frameworks have been applied across diverse domains, including advanced web research~\cite{erdogan2025plan, zhang2025webpilot, prasad2023adapt}, code generation~\cite{shinn2023reflexion, madaan2023self}, and mathematical problem-solving~\cite{chen2023agentverse}.

Similar to the dual-role frameworks proposed in various works~\cite{prasad2023adapt, erdogan2025plan}, our approach employs a multi-agent structure comprising of a designated Reasoner-Planner Agent (RPA) and Proxy-Execution Agents (PEA) that carry out the RPA instructions. However, our contribution diverges by evaluating on multi-step problems demanding the use of diverse, structured tools and different knowledge bases, reflecting the complexity of enterprise tasks.

\paragraph{\textbf{Large Reasoning Models}}
Large Reasoning Models (LRMs) have emerged as a pivotal advancement in language model capabilities, distinguished by their ability to generate explicit, intermediate steps of reasoning.
They are often referred as "\textit{rationales}" or "\textit{thoughts}" that deconstruct and solve complex, multi-step problems. This process is analogous to deliberative human cognitive faculties, such as systematic search and reflective analysis~\cite{qin2024o1, xu2025towards}.
Several distinct methodologies have been developed to elicit and enhance these reasoning capabilities such as Chain-of-Thought (CoT)~\cite{zelikman2024star} or Monte Carlo Tree Search (MCTS)~\cite{zhao2024marco, wu2024beyond}. Subsequent research has focused on directly embedding reasoning abilities into the model's parameters through Supervised Fine-Tuning~\cite{hosseini2024v} or through reinforcement learning~\cite{uesato2022solving, wang2023math}.
The efficacy of these methods is demonstrated by a growing number of powerful LRMs. This includes open-weight models~\cite{yang2025qwen3, liu2024deepseek, hosseini2024v}, and closed models~\cite{jaech2024openai}. The integration of explicit reasoning has proven to be a critical factor in advancing performance across a wide range of applications, including medical reasoning, math problems and information retrieval~\cite{huang2025o1, li2025search, uesato2022solving}.

Our work extends the strategy proposed in~\citealt{li2025search}. While their approach uses a Retrieval-Augmented Generation (RAG) system to query a static knowledge base, our method introduces a more dynamic, interactive framework. We replace the RAG component with multiple dedicated Proxy-Execution Agent that interfaces with a variety of external tools. The primary Reasoner-Planner Agent delegates sub-questions to these proxies, which then execute the necessary actions to find an answer.
\begin{figure*}[!t]
    \centering
    \includegraphics[width=1.0\linewidth]{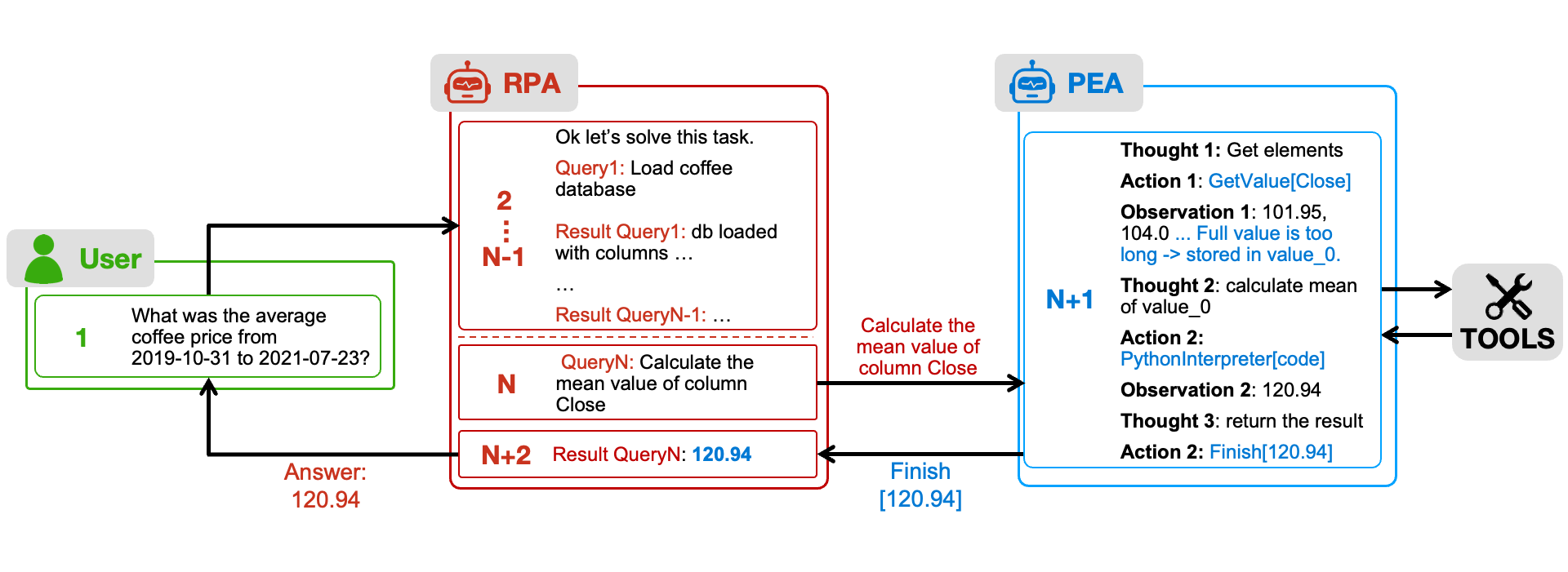}
    \caption{A PEA receives the sub-question from the RPA and tries to solve it by interacting with tools using the ReAct approach, then returns the result to the RPA.}
    \label{fig:pea-react-example}
\end{figure*}
\section{Methodology}
The nature of modern enterprise domains presents a significant challenge: agents must solve complex tasks that require orchestrating a multitude of sub-steps across various external tools. Effective task completion demands two distinct parts:
\begin{itemize}
   \item  High-Level Strategies: Developing a coherent, multi-step plan to achieve the overall goal.
   \item  Low-Level Strategies: Executing multiple precise tool calls, which involves identifying the correct service, managing specific input parameters, and following to the required syntax.
\end{itemize}
Furthermore, a robust agent must handle dynamic environments: it must reason critically over the tool's response and, consequently, determine the next course of action or re-plan entirely if an execution step fails.

When a single model is tasked with simultaneously managing both the planning and the execution of multiple, low-level tool actions, it faces a severe cognitive load, resulting in suboptimal decisions or inconsistent task completion~\cite{erdogan2025plan}.

To mitigate this complexity and the resulting context overload, we adopt a divide-and-conquer strategy, structurally dividing the problem into two distinct, roles: planning and execution. We introduce a multi-agent architecture composed of a Reasoner-Planner Agent (RPA), responsible for all high-level reasoning and planning, and multiple Proxy-Execution Agent (PEA) (Figure~\ref{fig:pea-react-example}).
The PEAs receive abstract instructions (e.g., "load database," "calculate mean values") from the RPA and they then convert these instructions into specific low level tool commands (loading databases, generating queries or executing code). This design is crucial: all the complexity and potential errors associated with tool usage are offloaded to the PEA and never enter the RPA's context window. This allows the RPA to maintain a clean, uncluttered context, enabling it to reason more freely and stably.
\paragraph{Reasoner-Planner Agent (RPA):}
The Reasoner-Planner Agent serves as the architecture's strategic core, responsible for all high-level adaptive planning. The RPA's primary function is to transform a complex, natural-language user request into a much simpler and actionable sub-questions.
The RPA communicates its plan to the execution component using a specific protocol. Each sub-question formulated during the "thinking" process is encapsulated between the tags \textcolor{red}{\texttt{<|begin\_search\_query|>}} and \textcolor{red}{\texttt{<|end\_search\_query|>}}.
Upon receiving the execution outcome, the RPA reads the result, which is delimited by the tags \textcolor{blue}{\texttt{<|begin\_search\_result|>}} and \textcolor{blue}{\texttt{<|end\_search\_result|>}}. It then leverages its reasoning capabilities to critically evaluate this feedback:
\begin{itemize}
    \item \textbf{Success}: If the execution result aligns with the plan's expectations, the RPA immediately proceeds to generate the next sub-question in the sequence.
    \item \textbf{Failure}: If the execution fails or produces an unexpected output, the RPA initiates a self-correction mechanism. It reasons to diagnose the probable cause of the failure and dynamically formulates a new corrective step or an entirely alternative trajectory to bypass the error and ensure continued progress toward the final objective.
\end{itemize}
    
\paragraph{Proxy-Execution Agent (PEA):}
The Proxy-Execution Agent (PEA) is dedicated to tool interaction. It functions as a reliable intermediary, receiving specific sub-questions from the RPA and translating them into the precise, low-level tool calls. This translation includes identifying the correct tool, specifying the necessary parameters, and adhering to the required syntax. When the sub-question is completed, the PEA returns the result back to the RPA.
To ensure high performance on single agent execution, the PEA employs the ReAct agent architecture \cite{yao2023react}. This approach allows the agent to iteratively refine its tool-use process through internal thought, tool invocation, and observation of the tool's response.
\paragraph{Context Management}
In an enterprise domain, the PEA frequently interacts with and extracts informations from large documents and databases (e.g., CSV, SQL, or txt files). This poses a significant challenge: a tool call that returns excessive data can severely degrade agent performance by overwhelming the context window (a critical constraint for open-weight LLMs) and potentially distracting the model from its primary goal. This is particularly pronounced when handling tabular data; to perform operations such as calculating a mean value, the agent does not require the full dataset. Instead, it only needs to inspect representative sample to generate the Python code necessary for analysis. Therefore, to solve this critical issue, we implement a specialized memory optimization strategy.
This mechanism operates on a defined token threshold, $T$. If the output returned by SQL query or a filtered CSV file, exceeds this limit, we adopt the following procedure:
\begin{itemize}
    \item Only the first $T$ tokens of the output are immediately injected into the agent's context window. This provides the agent with a crucial initial preview of the data structure and content.
    \item The original text is saved to a newly created temporary variable stored within the agent's execution environment.
\end{itemize}
With this partial preview, the agent is then aware that analysing the complete information stored in the variable (e.g., calculating the mean) will necessitate the use of the Python execution tool.
Even if the RPA's request was only to retrieve information, the PEA returns the data preview and the variable name. It explicitly warns the RPA that the entire output volume requires excessive context consumption, necessitating further analysis via the Python tool.
This simple technique is critical as it acts as a safeguard against context consumption, ensuring the agent's context remains clean and focused to avoid trajectory deviation.

\section{Experimental Setup}

\subsection{Dataset}
We evaluate our architecture on the \textbf{ToolQA} dataset \cite{zhuang2023toolqa}, a comprehensive benchmark for easy and complex question answering task that require the usage of various tools. The dataset provides complex, multi-step problems across eight domains, categorized into easy and hard difficulties based on the required number of tool interactions and reasoning complexity. The agent is provided with a toolkit of 13 functions, including various utilities such as text retrieval, database queries, code interpretation, and mathematical computation.
Specifically, our evaluation was performed on both the \textbf{easy} and \textbf{hard} subsets of five key domains: \textbf{Airbnb}, \textbf{Flight}, \textbf{Coffee}, \textbf{Scirex}, and \textbf{Yelp}.

\subsection{Baseline Agent Architectures}
We compare our architecture against two established baseline architectures:
\begin{itemize}
\item \textbf{ReAct} \cite{yao2023react}:
The first baseline employs the standard ReAct methodology. In this framework, the agent resolves tasks through a continuous loop defined by the sequential steps: Think (formulating the next step of the plan), Action (applying the generated command to the environment), and Observe (integrating the environmental feedback). The planning and execution responsibilities are unified within this single component. This approach shows impressive performance on various decision-making tasks such as HotPotQA~\cite{yang2018hotpotqa} and ALFWorld~\cite{shridhar2020alfworld}.
\item \textbf{Reflexion} \cite{shinn2023reflexion}
The second baseline is the Reflexion agent, which augments the ReAct cycle with self-correction capabilities. This agent incorporates a separate stage for evaluating its past performance and a subsequent self-reflection mechanism. This allows the model to generate advices that can be used to correct the trajectory of a failed execution. This approach is particularly effective on various tasks such as decision-making \cite{yang2018hotpotqa} and programming tasks \cite{chen2021evaluating}.
\end{itemize}

\subsection{Prompts}
In order to conduct a test our approach against a React and Reflexion agent, we used the examples proposed in the original work by~\citealt{zhuang2023toolqa}.
We reuse their React prompt but adapt it structurally to align with the requirements of our architecture.
Specifically for the React examples, each complete execution is divided into $n$ sequential steps. First, for every step, we explicitly define the potential question or query that the RPA will send to the PEA. Second, for each of these potential queries, we also define the precise sequence of React steps (Think, Act, Observate) that the PEA should execute to successfully resolve the query.

The RPA uses the same prompt and example set across all domains. The PEA, however, adapts its few-shot examples based on the domain-specific tools managed, while keeping the underlying system prompt constant.
In our experimental evaluation, we tested a unified PEA configuration tasked with managing the entire suite of available tools. Consequently, its prompt was populated with the complete set of few-shot examples.
For complete transparency, the full collection of prompts employed during this evaluation is provided in the Appendix.

\begin{table*}[!ht]
\centering
\scriptsize
\setlength{\tabcolsep}{4pt}

\begin{minipage}[t]{.45\textwidth}
\centering
\caption{$Accuracy$ on easy Benchmarks.}
\label{tab:easy-benchmark}
\begin{tabular}{ll*{5}{S[table-format=1.2]}}
\toprule
\multicolumn{2}{c}{Method} &
\multicolumn{1}{c}{Yelp} & \multicolumn{1}{c}{SciREX (\textcolor{blue}{P})} & 
\multicolumn{1}{c}{Flight (\textcolor{blue}{P})} & \multicolumn{1}{c}{Airbnb} &
\multicolumn{1}{c}{Coffee (\textcolor{blue}{P})} \\
\cmidrule(lr){1-2}
Agent Approach & Model & \multicolumn{5}{c}{} \\
\midrule
React      & gpt-oss-120b & \textbf{0.90} & \textbf{0.17} & \textbf{0.76} & 0.73 & 0.78 \\
Reflexion  & gpt-oss-120b & 0.87 & 0.07 & 0.48 & 0.73 & \textbf{0.80} \\
RP-ReAct   & gpt-oss-120b & 0.53 & 0.09 & 0.50 &  \textbf{0.89} & 0.78 \\
\addlinespace
React      & Qwen3-32B    & \textbf{0.76} & \textbf{0.09} & \textbf{0.54} & \textbf{0.85} & 0.72\\
Reflexion  & Qwen3-32B    & 0.39 & 0.04 & 0.43 & 0.28 & 0.74 \\
RP-ReAct   & Qwen3-32B    & 0.68 & 0.07 & 0.43 & 0.69 & \textbf{0.78}\\
\addlinespace
React      & gpt-oss-20b  & 0.15 & 0.04 & \textbf{0.11} & 0.11 & 0.05 \\
React-100  & gpt-oss-20b  &\text{0.22} & \text{0.04} & \text{0.12} & \text{0.18} & \text{0.07} \\
Reflexion & gpt-oss-20b  & 0.04 & 0.05 & 0.04 & 0.09 & 0.08 \\
RP-ReAct   & gpt-oss-20b  & \textbf{0.26} & \textbf{0.07}& 0.09 & \textbf{0.48} & \textbf{0.43} \\
\addlinespace
React      & Qwen3-14B    & \textbf{0.71} & \textbf{0.08} & \textbf{0.49} & \textbf{0.76} & \textbf{0.67} \\
Reflexion  & Qwen3-14B    & 0.43 & 0.02 & 0.23 & 0.30 & 0.49\\
RP-ReAct   & Qwen3-14B    & 0.61 & 0.05 & 0.37 & 0.72 & 0.65 \\
\addlinespace
\midrule
ReAct & DeepSeek-8B & 0.17 & \llap{$\approx\mkern-15mu$}0.00 & \llap{$\approx\mkern-8mu$}0.00 & \textbf{0.33} & 0.09 \\
RP-ReAct & DeepSeek-8B & \textbf{0.30} & \llap{$\approx\mkern-15mu$}0.00 & \llap{$\approx\mkern-8mu$}0.00 & 0.20 & \textbf{0.28}\\
\addlinespace
ReAct & DeepSeek-7B & 0.11 & \llap{$\approx\mkern-15mu$}0.00 & \llap{$\approx\mkern-8mu$}0.00 & 0.04 & \textbf{0.21}\\
RP-ReAct & DeepSeek-7B  & \textbf{0.11} & \llap{$\approx\mkern-15mu$}0.00 & \llap{$\approx\mkern-8mu$}0.00 & \textbf{0.04} & 0.06 \\
\bottomrule
\end{tabular}
\end{minipage}
\hfill 
\begin{minipage}[t]{.45\textwidth}
\centering
\caption{$Accuracy$ on hard Benchmarks.}
\label{tab:hard-benchmark}
\begin{tabular}{ll*{5}{S[table-format=1.2]}}
\toprule
\multicolumn{2}{c}{Method} &
\multicolumn{1}{c}{Yelp (\textcolor{blue}{P})} & \multicolumn{1}{c}{SciREX} & 
\multicolumn{1}{c}{Flight (\textcolor{blue}{P})} & \multicolumn{1}{c}{Airbnb} &
\multicolumn{1}{c}{Coffee} \\
\cmidrule(lr){1-2}
Agent Approach & Model & \multicolumn{5}{c}{} \\
\midrule
React      & gpt-oss-120b & \textbf{0.63}  & 0.14 & \textbf{0.22}  & 0.32 & 0.11 \\
Reflexion  & gpt-oss-120b  & 0.43 & 0.10 & 0.20  & 0.17 & 0.07 \\
RP-ReAct   & gpt-oss-120b  & 0.37  & \textbf{0.26} & 0.09 & \textbf{0.38} & \textbf{0.23} \\
\addlinespace
React      & Qwen3-32B    & \textbf{0.51} & 0.14 & \textbf{0.14} & 0.24 & 0.10 \\
Reflexion  & Qwen3-32B    & 0.29 & \textbf{0.33} & 0.10 & 0.16 & 0.14 \\
RP-ReAct   & Qwen3-32B    & 0.27 & 0.20 & 0.13 & \textbf{0.27} & \textbf{0.18} \\
\addlinespace
React      & gpt-oss-20b  & 0.02 & 0.13 & 0.04 & 0.10 & 0.13 \\
React-100  & gpt-oss-20b  & \text{0.06} & \text{0.13} & \text{0.07} & \text{0.14}& \text{0.29} \\
Reflexion  & gpt-oss-20b   & 0.06 & 0.13 & 0.02  & 0.06 &  0.07 \\
RP-ReAct   & gpt-oss-20b  & \textbf{0.16}  & 0.13 & \textbf{0.08}  & \textbf{0.28} & \textbf{0.44} \\
\addlinespace
React      & Qwen3-14B    & \textbf{0.47} & 0.14 & 0.13 & \textbf{0.23} & 0.03 \\
Reflexion  & Qwen3-14B     & 0.24 & 0.07 & 0.08 & 0.06 & 0.07 \\
RP-ReAct   & Qwen3-14B     & 0.21 & \textbf{0.17} & \textbf{0.18} & 0.15 & \textbf{0.20} \\
\addlinespace
\midrule
ReAct & DeepSeek-8B & \textbf{0.10} & \llap{$\approx\mkern-6mu$}0.00 & \llap{$\approx\mkern-8mu$}0.00 & \textbf{0.11} & 0.03 \\
RP-ReAct & DeepSeek-8B  & 0.03 & \llap{$\approx\mkern-6mu$}0.00 & \llap{$\approx\mkern-8mu$}0.00 & 0.07 & \textbf{0.04}\\
\addlinespace
ReAct & DeepSeek-7B & \textbf{0.10}& \llap{$\approx\mkern-6mu$}0.00 & \llap{$\approx\mkern-8mu$}0.00 &  \textbf{0.07} & \textbf{0.09} \\
RP-ReAct & DeepSeek-7B & 0.01 & \llap{$\approx\mkern-6mu$}0.00 & \llap{$\approx\mkern-8mu$}0.00 & 0.02 & 0.01 \\
\bottomrule
\end{tabular}
\end{minipage}
\end{table*}

\subsection{Models}
Given the enterprise necessity of using open-weight models for data privacy, we conducted a comprehensive evaluation of our architecture across different scales by using a selection of six open-weight Large and Small Reasoning Models.
Our selection includes models of different sizes to assess performance with both resource-constrained and larger foundation models.
Specifically, we tested gpt-oss 20B and 120B \cite{agarwal2025gpt}, Qwen3 14B and 32B \cite{yang2025qwen3}, DeepSeek-R1-Distill-Qwen-7B and DeepSeek-R1-Distill-Llama-8B \cite{guo2025deepseek}.

\subsection{Evaluation Metrics}
We evaluate the performance of the models across each benchmark and architecture using a comprehensive set of metrics.
Our primary metrics is Accuracy ($Acc$), used to quantify the model's performance for each agent approach and domain of use. Then we used Standard Deviation ($Std$), which serves to assess the stability and consistency of each architecture across models.
To provide a more holistic assessment that captures the trade-off between best performance and stability, we also include several combined metrics derived from recent works \cite{mizrahi2024state, magnini2025leaderboard}: \textit{Saturation}, and \textit{Combined Performance Score}.

\subsubsection{Saturation}
Considering $M$ as the set of models, we evaluate an architecture $a \in A$ with the following metrics:
$$Sat_a=1-(MaxAcc_a-AverageAcc_a)$$
Where:
\[
{MaxAcc_a}=\max_{m \in M} Accuracy(m, a)
\]
\[
{AverageAcc_a}
= \frac{1}{|M|}\sum_{m \in M} Accuracy(m, a)
\]
\subsubsection{Combined Performance Score (CPS)}
This score integrates both stability (robustness) and best observed performance:
$$CPS_a = Sat_{a} \cdot MaxAcc_a$$
By considering both stability and accuracy, this metric can summarize how well an approach consistently performs well on different tasks and models. 

\subsection{Computational Resources and Hyperparameters}
We run our experiments on Leonardo HPC with 4x NVIDIA A100 GPU 64GB, 128 GB RAM and 32-core Intel Xeon Platinum 8358 CPU.
We set the temperature to 0.6 and TopP to 1.0. As we are experimenting this architecture for the first time, for simplicity's sake we only test our approach in the simple configuration of 1 RPA and 1 PEA. 
We set the threshold $T=100$. For ReAct we set the number of steps $N=20$; For Reflexion we set $N=20$ and max 3 self reflections. For RP-ReAct agents we set $N=10$.
\begin{table*}[!t]
\centering
\caption{$Mean$ ($\uparrow$), $Standard Deviation$ ($\downarrow$) and $Combined Performance Score$ ($\uparrow$) of each Agent Approach}
\label{tab:main-benchmarks}
\scriptsize
\setlength{\tabcolsep}{4pt}

\begin{subtable}[t]{\textwidth}
    \centering
    \caption{easy Benchmarks}
    \label{tab:benchmark-a}
    \begin{tabular}{l*{5}{S[table-format=1.2]S[table-format=1.2]S[table-format=1.2]}}
    \toprule
    \multicolumn{1}{c}{} & \multicolumn{3}{c}{\textbf{Yelp}} & \multicolumn{3}{c}{\textbf{SciREX} (\textcolor{blue}{P})} & \multicolumn{3}{c}{\textbf{Flight} (\textcolor{blue}{P})} & \multicolumn{3}{c}{\textbf{Airbnb}} & \multicolumn{3}{c}{\textbf{Coffee} (\textcolor{blue}{P})} \\
    \cmidrule(lr){2-4} \cmidrule(lr){5-7} \cmidrule(lr){8-10} \cmidrule(lr){11-13} \cmidrule(lr){14-16}
    \multicolumn{1}{c}{Agent Approach} & {Mean} & {Std} & {CPS} & {Mean} & {Std} & {CPS} & {Mean} & {Std} & {CPS} & {Mean} & {Std} & {CPS} & {Mean} & {Std} & {CPS} \\
    \midrule
    React & \textbf{0.63} & 0.32 & \textbf{0.65} & \textbf{0.09}  & 0.05& \textbf{0.15} & \textbf{0.47} & 0.27 & \textbf{0.54}& 0.61 &  0.33 & 0.64&0.55 &0.07& 0.60  \\
    Reflexion  & 0.43 & 0.34& 0.48& 0.04 & 0.02& 0.06& 0.29 & 0.20 & 0.39& 0.35 & 0.32& 0.45& 0.52 & 0.07 & 0.58 \\
    RP-ReAct    & 0.52 & \textbf{0.18}& 0.57& 0.07 & \textbf{0.01}& 0.08& 0.34 & \textbf{0.17} & 0.42& \textbf{0.69 } & \textbf{0.16}& \textbf{0.71}& \textbf{0.66} & \textbf{0.04}& \textbf{0.68} \\
    \bottomrule
    \end{tabular}
\end{subtable}

\vspace{1em} 

\begin{subtable}[t]{\textwidth}
    \centering
    \caption{hard Benchmarks}
    \label{tab:benchmark-h}
    \begin{tabular}{l*{5}{S[table-format=1.2]S[table-format=1.2]S[table-format=1.2]}}
    \toprule
    \multicolumn{1}{c}{} & \multicolumn{3}{c}{\textbf{Yelp} (\textcolor{blue}{P})} & \multicolumn{3}{c}{\textbf{SciREX}} & \multicolumn{3}{c}{\textbf{Flight} (\textcolor{blue}{P})} & \multicolumn{3}{c}{\textbf{Airbnb}} & \multicolumn{3}{c}{\textbf{Coffee}} \\
    \cmidrule(lr){2-4} \cmidrule(lr){5-7} \cmidrule(lr){8-10} \cmidrule(lr){11-13} \cmidrule(lr){14-16}
    \multicolumn{1}{c}{Agent Approach} & {Mean} & {Std} & {CPS} & {Mean} & {Std} & {CPS} & {Mean} & {Std} & {CPS} & {Mean} & {Std} & {CPS} & {Mean} & {Std} & {CPS} \\
    \midrule
    React   & \textbf{0.40}& 0.26& \textbf{0.49}& 0.13 & \textbf{0.00}& 0.14& \textbf{0.13} & 0.07 & 0.16& 0.22 &0.09& 0.28&0.08 & 0.04& 0.12 \\
    Reflexion   & 0.25 & 0.15& 0.35& 0.15 & 0.11& \textbf{0.27}& 0.12 & 0.07 & 0.18& 0.11 & \textbf{0.06}& 0.16& 0.09 & \textbf{0.03} & 0.13  \\
    RP-ReAct   & 0.25 & \textbf{0.09}& 0.32 & \textbf{0.19} & 0.05& 0.24& 0.10 & \textbf{0.04} & \textbf{0.20}& \textbf{0.26} & 0.09& \textbf{0.33}& \textbf{0.27} & 0.12 & \textbf{0.36}  \\
    \bottomrule
    \end{tabular}
\end{subtable}

\end{table*}
\section{Results}

\subsection{Performance Analysis}
In Table \ref{tab:easy-benchmark} and \ref{tab:hard-benchmark}  we report the accuracy for the easy and hard benchmarks respectively.
As expected, the easier benchmarks garner higher scores compared to the harder ones. 
In our analysis, we will focus on the comparison between React and RP-React because Reflexion shows the worst performance overall in basically all tasks.
Conducting a qualitative and quantitative analysis on the approaches' outputs, we noticed that React works well when the required tasks can be solved in few steps (easy benchmark). 
This trend does not carry on to the harder benchmarks. In fact, our proposed approach produces better results in most domains. 

Looking to the agent trajectories, this counter-intuitive outcome is due to two factors:
\begin{itemize}
    \item For easy tasks, the inherent planning overhead in RP-ReAct often disrupts the optimal trajectory. The Reasoner-Planner Agent (RPA) may introduce redundant actions (such as repeated database loading) or initiate unnecessary verifications and re-planning steps. The simpler, monolithic ReAct agent, is more effective at choosing the most direct steps for straightforward goals.
    \item For hard tasks requiring numerous steps, diverse tool interactions, and complex reasoning, the explicit separation of planning and execution is crucial. Our approach is significantly more effective at preventing the RPA from losing its trajectory. The monolithic ReAct approach, in contrast, tends to fail because its context window becomes overwhelmed by large tool outputs or frequent wrong tool calls, leading the agent to deviate from the correct path. RP-ReAct maintains stability by separating the planner from this low-level execution noise. 
\end{itemize}

As we can see in Table \ref{tab:easy-benchmark} and \ref{tab:hard-benchmark} (where \textcolor{blue}{P} marks specific domain examples included in the prompt), RP-ReAct demonstrates a significantly more consistent performance and superior generalization ability, even when specific examples are omitted from the prompt. 
The key to RP-ReAct's generalization lies in its structural division. Where the monolithic ReAct approach relies heavily on rigid, end-to-end example trajectories tied to complex questions, our explicit separation enables a greater generalization of low-level tool usage across diverse domains. For instance, to learn the necessary steps for a fundamental tool action, ReAct requires multiple, complete task examples that contain that specific action. In contrast, the RP-ReAct PEA only needs one dedicated, abstract example where the instruction is simply to perform that standardized tool operation. This crucial abstraction effectively decouples the tool usage logic from the overall task logic. Consequently, knowledge of "how to use Tool X" is learned with one sample, significantly improving sample efficiency and allowing the PEA to execute the action correctly regardless of the broader task domain.

\subsection{Smaller Models}
Due to the low performance of Reflexion, we limited our analysis of the smaller models to a comparison of our approach only against ReAct.
In general, as we can see at the bottom of Tables \ref{tab:easy-benchmark} and \ref{tab:hard-benchmark}, smaller models (with less than 10B parameters) are still unable to solve tasks consistently regardless of the approach. 
We did not report results for SiREX and Flight for both the easy and hard versions as their performance approached 0.
Also, as we can notice that in the hard benchmarks, the results do not surpass the 0.11 threshold, indicating the impossibility to evaluate our approaches with models with such small parameters.
Qualitative analysis attributes these failures primarily to two factors:
\begin{itemize} 
\item Premature Response: The PEA tends to generate immediate answers rather than delegating the necessary sub-tasks to the RPA, effectively ignoring prompt constraints. 
\item Trajectory Deviation: The RPA struggles to maintain a valid execution trajectory, often selecting incorrect tool parameters or step sequences (e.g. filtering the db before loading), which causes the agent to hit the step limit before solving the task.
\end{itemize}
We want to stress the fact that these models have not been subject to post-training in any sort of way. 
Our approach has shown better results for specific tasks but does not translate into higher performance across domains.

\subsection{Analysis of Trajectory Against Execution Step Limits}
As discussed in the previous Sections, our approach demonstrates superior generalisation capabilities across diverse problem domains. This advantage is particularly evident in the hard domain, which is characterized by a high number of requisite sequential steps to achieve the final goal. The ReAct methodology, conversely, is more susceptible to execution drift and error propagation when faced with a high volume of necessary tool interactions, frequently leading to either an incorrect final output or premature termination (reaching the maximum available step limit).
In our settings, the React agent has a maximum step limit of 20 steps and the RP-ReAct has a step limit of 10 for the RPA and 10 for the PEA (each time a new question is received), meaning that in the worst case scenario it can reach 100 steps.
So, to further confirm that React’s performance degradation is due to trajectory failure rather than merely a limited number of execution steps, we conducted a targeted investigation.
We identified all questions where RP-ReAct found the solution but the React agent failed by reaching the maximum step limit. We then re-executed the React agent on this subset, setting its maximum step limit with the same for RP-ReAct (100). 
We did this test with gpt-oss-20b where we encountered the highest number of runs where the React agent reached the maximum number of steps.
As we can see in Tables \ref{tab:easy-benchmark} and \ref{tab:hard-benchmark} (React-100), the results showed that performance increased by an average of only 4.8\%. In the remaining cases, the React agent either followed a wrong trajectory, produced an incorrect answer, or reached the new 100-step limit again. This confirms that the structured planning of RP-ReAct outperforms the standard React approach, validating our hypothesis.
This highlights that simply increasing the number of steps is not beneficial for reaching the goal; if the model commits to an incorrect trajectory, it will fail to find the solution regardless. This shows that the ReAct approach lacks the robustness required for complex question-answering tasks as it seems to hit a plateau in performance after a limited number of steps.

\subsection{Evaluating Robustness and Stability}
In the agent domain, the majority of works do not test their proposed approach on different open-weight models, often focusing on a single big model or closed source models \cite{wu2024avatar, yao2023react}. To address this, we propose an analysis that explicitly tests the robustness and stability of the agent approaches across a diverse set of models with varying parameter sizes: gpt-oss (20B, 120B) and Qwen3 (14B, 32B).
As explained in the previous Section, we conduct the analysis separating on the easy (Table \ref{tab:benchmark-a}) and the hard benchmark (Table \ref{tab:benchmark-h}).
\subsubsection{Analysis of Mean Accuracy}
As a preliminary measure, we calculated the mean accuracy for each Agent approach, averaging the performance across all four models for each specific benchmark.
The results confirm our preceding analysis of the different performance based on the difficulty of the task. ReAct consistently shows higher mean accuracy in easy tasks (Yelp, SciREX, and Flight), particularly in scenarios where task resolution examples are included within the prompt (e.g., easy SciREX, easy Flight and hard Yelp, hard Flight). Conversely, RP-ReAct exhibits superior mean accuracy on hard tasks, confirming its enhanced ability to generalize in complex, multi-step tasks that lack in-context examples.
\subsubsection{Stability Metrics and the Performance Trade-Off}
The stability indices offer a more critical insight. We observed that RP-ReAct is significantly more stable than both ReAct and Reflexion. This is quantitatively demonstrated by the standard deviation, where our approach shows noticeable lower performance variability across the different models in the majority of both easy and hard tasks. 
While Reflexion achieves a low standard deviation in several cases, this apparent stability is critically undermined by its significantly low accuracy compared to the other two methodologies, posing challenges for robust real-world application.

To perform a comprehensive assessment that balances both high performance and stability, we used the Combined Performance Score (CPS). The results show that RP-ReAct achieves the optimal trade-off between stability and performance, particularly for the challenging hard benchmarks. ReAct performs better with the CPS primarily for the easy tasks guided by in-prompt examples, and one example-guided hard task.
In conclusion, this combined analysis validates our central finding: the RP-ReAct framework ensures superior generalization over both ReAct and Reflexion. It enables various models to achieve much more robust performance across diverse domains while simultaneously showing better stability and reliability across different model scales and architectures.

\section{Conclusions}
In this study we introduced RP-React, a novel multi-agent architecture designed to enhance the performance and reliability of autonomous agents in enterprise tasks where multiple agents need to be orchestrated to provide support to workers. By decoupling the cognitive load into a dedicated Reasoner Planner Agent and Proxy Executor Agents, RP-React moves beyond the limitations of conventional single-agent React approaches.
We tested our architecture on five domains of ToolQA dataset (both easy and hard) and 13 available tools.

The empirical results demonstrate that RP-React exhibits better performance against the state-of-the-art, particularly in hard tasks that necessitate intensive reasoning and a high number of sequential tool interactions. We also confirm it by running the step-limit analysis, which demonstrated that merely increasing the maximum number of execution steps provides negligible benefit to the standard ReAct approach. Crucially, the functional separation of planning from action execution significantly improves the agent's generalization ability that do not necessary need inter-domain example of executions.
We observe that smaller models (with less than 10B parameters) are still too weak to solve the tasks of this benchmark consistently, regardless of the approach used.
Finally, the architecture showcases high stability and robustness across various open-weight models in most of the tested domains. This suggests practical viability for the enterprise sector, particularly for organizations deploying open-weight models on resource-constrained hardware.

\section{Limitations and Future Works}
This exploratory work presents several limitations that serve as directions for future research:
\begin{itemize}
\item To fully assess generalizability, we plan to expand our evaluation to other prominent complex reasoning benchmarks, including OfficeBench \cite{wang2024officebench} and Mint \cite{wang2023mint}, employing an increased number of PEAs for testing.

\item The reported performance reflects the baseline capability of our agentic approach, as we did not employ post-training optimization techniques (e.g., SFT or RL). We expect that further post-training specific for the RPA and the PEA can improve their performance reducing redundant steps and wrong replanning actions.

\item Our current analysis of context management offers a preliminary view of token savings. A more comprehensive investigation is needed to quantify the optimal performance characteristics, specifically by analysing the effect of various $T$ thresholds or integrating a token summarization mechanisms.
\item Our experiments used a static temperature of 0.6. Future iterations will explore agent-specific temperature tuning, such as lowering the RPA's temperature to 0.0 to prioritize deterministic outputs over creativity.
\end{itemize}

\section{Acknowledgments}
We acknowledge ISCRA for awarding this project access to the LEONARDO supercomputer, owned by the EuroHPC Joint Undertaking, hosted by CINECA (Italy) .
Fabio Ciravegna was partially funded  by project ICOS "Towards a functional continuum operating system", funded by the European Union’s HORIZON research and innovation programme under grant agreement No 101070177.
\bibliography{aaai2026}

\section{Appendix}
Here we include the prompts given to the agents.
\paragraph{React-style prompt}
\begin{quote}
\small\ttfamily
Solve a question answering task with interleaving Thought, Action, Observation steps. Thought can reason about the current situation, and Action can be 13 types: \\
(1) Calculate[formula], which calculates the formula and returns the result.\\
(2) RetrieveAgenda[keyword], which retrieves the agenda related to keyword.\\
(3) RetrieveScirex[keyword], which retrieves machine learning papers' paragraphs related to keyword.\\
(4) LoadDB[DBName], which loads the database DBName and returns the database. The DBName can be one of the following: flights/coffee/airbnb/yelp.\\
(5) FilterDB[condition], which filters the database DBName by the column column\_name the relation (e.g., =, >, etc.) and the value value, and returns the filtered database.\\
(6) GetValue[column\_name], which returns the value of the column column\_name in the database DBName.\\
(7) LoadGraph[GraphName], which loads the graph GraphName and returns the graph. The GraphName can be one of the following: PaperNet/AuthorNet.\\
(8) NeighbourCheck[GraphName, Node], which lists the neighbours of the node Node in the graph GraphName and returns the neighbours.\\
(9) NodeCheck[GraphName, Node], which returns the detailed attribute information of Node.\\
(10) EdgeCheck[GraphName, Node1, Node2], which returns the detailed attribute information of the edge between Node1 and Node2.\\
(11) SQLInterpreter[SQL], which interprets the SQL query SQL and returns the result.\\
(12) PythonInterpreter[Python], which interprets the Python code Python and returns the result.\\
(13) Finish[answer], which returns the answer and finishes the task.\\[0.5ex]
You may take as many steps as necessary.\\
It is extremely important that you conclude each Thought with ``.''\\[0.5ex]
Here are some examples:\\
\{examples\}\\
(END OF EXAMPLES)\\[0.5ex]
Question: \{question\}\{scratchpad\}
\end{quote}

\paragraph{Reflexion evaluator prompt}
\begin{quote}
\small\ttfamily
You are an agent EVALUATOR. your job is to evaluating the trajectory of the agent that tried to solve a question.
Giving the question and the agent trajectory, you ONLY have to output:
- [SUCCESS] if you think that the agent solved the question
- [FAILURE] if you think that the agent did not solve the question
It is extremely important that you put your verdict inside [], for example [SUCCESS] or [FAILURE].
Give me only the verdict, do not write anything else.
This is the question: {question}
This is the agent trajectory: {trajectory}
This is your verdict ([SUCCESS] or [FAILURE]):
\end{quote}

\paragraph{Reflexion self-refine prompt}
\begin{quote}\small\ttfamily
You are an advanced REASONER agent that can improve the agent trajectory based on self reflection. You will be given a previous trial in which the agent were given access to the aviable tools and a question to answer . The agent were unsuccesfull in answering the question either because it failed to solve the task or give the answer, or it used up your set number of reasoning steps. In a few sentences, diagnose a possible reason for failure and devise a new, concise, high level plan that aims to mitigate the same failure. Use complete sentences and do not write more than 3 lines.  

- Aviable tools:\\
(1) Calculate[formula], which calculates the formula and returns the result.\\
(2) RetrieveAgenda[keyword], which retrieves the agenda related to keyword.\\
(3) RetrieveScirex[keyword], which retrieves machine learning papers' paragraphs related to keyword.\\
(4) LoadDB[DBName], which loads the database DBName and returns the database. The DBName can be one of the following: flights/coffee/airbnb/yelp.\\
(5) FilterDB[condition], which filters the database DBName by the column column\_name, the relation (e.g., =, >, etc.) and the value value, and returns the filtered database.\\
(6) GetValue[column\_name], which returns the value of the column column\_name in the database DBName.\\
(7) LoadGraph[GraphName], which loads the graph GraphName and returns the graph. The GraphName can be one of the following: PaperNet/AuthorNet.\\
(8) NeighbourCheck[GraphName, Node], which lists the neighbours of the node Node in the graph GraphName and returns the neighbours.\\
(9) NodeCheck[GraphName, Node], which returns the detailed attribute information of Node.\\
(10) EdgeCheck[GraphName, Node1, Node2], which returns the detailed attribute information of the edge between Node1 and Node2.\\
(11) SQLInterpreter[SQL], which interprets the SQL query SQL and returns the result.\\
(12) PythonInterpreter[Python], which interprets the Python code Python and returns the result.\\
(13) Finish[answer], which returns the answer and finishes the task.\\

It is extremely important that you conclude your sentence with ``.''\\
Use the thinking tags \textless think\textgreater{} ... \textless/think\textgreater{} to reason the output Thought or Action to give and then output the final reflection.\\
Do not write Thought or Action tags.\\
Previous trial to reflect on:\\
- Question: \{question\}\\
- Trajectory: \{trajectory\}\\
- Previous reflections: \{prev\_reflections\}\\[0.5ex]
REFLECTION:
\end{quote}

\paragraph{RPA prompt}
\begin{quote}
\small\ttfamily
You are a reasoning assistant agent with the ability to perform high-level questions to help you answer the user's question accurately. These questions will be sent to an *Executor* agent, which will translate them into specific tool calls and return the results. You will then use these results to formulate your next question or to provide the final answer.
1.  HOW TO ASK :
To perform a search: write <|begin\_search\_query|> your query here <|end\_search\_query|>.
Example: <|begin\_search\_query|>Load the flights database<|end\_search\_query|>
Then, the *Executor* agent will convert the query into actionable tool calls and return the results in the format <|begin\_search\_result|> ...search results... <|end\_search\_result|>.
Inside the tag:
  -  Describe the desired action in plain English, e.g.  
    - "Load the flights database."  
    - "Filter the flights\_db table for flight DL82 on 2022-01-18."  
    - "Return the DepTime column of that row."  
  -  Mention input variables if the action needs them (e.g. write a python program that count all the substrings divided by a space in the variable value0).  
2.  REASON \& FINISH:
- You can repeat the search process multiple times if necessary. The maximum number of search attempts is
limited to {MAX\_SEARCH\_LIMIT}.
- Use the <think> </think> tags to reason on the question, the tool response and plan the next action.
- Always use use factual data returned by the *Executor* agent to give an answer.
- When you have the final answer, close the think block and output:
  <Finish> answer </Finish>
- Always check that the tool return the expected data, if not, rewrite your question.
- If between the returned results tags <|begin\_search\_result|><|end\_search\_result|> no value is returned then the *Executor* agent failed to use the tool, so you have to rewrite your question in a way that the *Executor* agent can use the tools correctly.
3.  AVAILABLE LOW-LEVEL TOOLS:
You can ask these questions to the *Executor* agent:
  - load a DB (flights / coffee / airbnb / yelp).
  - if the DB is succesfully loaded you can ask to filter some data from the DB by conditions using the columns of the DB.
  - get all the data inside a column of the current DB (if there is too much data to read the *Executor* will return a variable that you have to analyze asking to write a python program with that variable and explaining what the program should do).
  - calculate arithmetic operations like +, -, *, / on numbers or mean, sqr etc...
  - retrive informations from the Agenda or Scirex items by keyword.
  - find ML-paper paragraphs by keyword.
  - load a graph (PaperNet / AuthorNet). 
  - list neighbour nodes.  
  - node attributes.  
  - edge attributes. 
  - get the data in a sql db on these domains (flights / coffee / airbnb / yelp) that follow your contraints (if there is too much data to read the *Executor* will return a variable that you have to analyze asking to write a python program with that variable and explaining what the program should do).
  - run Python code on data that you provide or on variables that the *Executor* will return to you.
  - finish the reasoning and give the final answer with tags <Finish> answer </Finish>.
4.  RULES :
**Output template you MUST follow**
<think>
your step-by-step reasoning...
</think>
<|begin\_search\_query|>...question and variables if needed...<|end\_search\_query|> 
<|begin\_search\_result|>...output...<|

end\_search\_result>
...(repeat as needed)...
<think>
...your step-by-step reasoning...
</think>
<Finish>...concise answer only...</Finish>
5. EXAMPLES:
{examples}
(END OF EXAMPLES)
QUESTION: {question}
{scratchpad}
\end{quote}

\paragraph{PEA prompt}
\begin{quote}
\small\ttfamily
Solve a question answering task with interleaving Thought, Action, Observation steps. Thought can reason about the current situation, and Action can be 13 types: 
(1) Calculate[formula], which calculates the formula and returns the result.
(2) RetrieveAgenda[keyword], which retrieves the agenda related to keyword.
(3) RetrieveScirex[keyword], which retrieves machine learning papers' paragraphs related to keyword.
(4) LoadDB[DBName], which loads the database DBName and returns the database. The DBName can be one of the following: flights/coffee/airbnb/yelp.
(5) FilterDB[condition], which filters the database DBName by the column column\_name the relation (e.g., =, >, etc.) and the value value, and returns the filtered database.
(6) GetValue[column\_name], which returns the value of the column column\_name in the database DBName.
(7) LoadGraph[GraphName], which loads the graph GraphName and returns the graph. The GraphName can be one of the following: PaperNet/AuthorNet.
(8) NeighbourCheck[GraphName, Node], which lists the neighbours of the node Node in the graph GraphName and returns the neighbours. 
(9) NodeCheck[GraphName, Node], which returns the detailed attribute information of Node. 
(10) EdgeCheck[GraphName, Node1, Node2], which returns the detailed attribute information of the edge between Node1 and Node2. 
(11) SQLInterpreter[SQL], which interprets the SQL query SQL and returns the result.
(12) PythonInterpreter(variable1, variable2...)[Python], which interprets the Python code Python and returns the result.
(13) Finish[answer], which returns the answer and finishes the task.
You may take as many steps as necessary.
It is extremely important that you conclude each Thought with "."
If asked inside the question you can pass to the PythonInterpreter variables using '(variables)' to read, modify and analyze their content.
If the solution of the question is inside a variable return the name of the variable and a few inside elements, else return the solution.
Here are some examples:
{examples}
(END OF EXAMPLES)
Previous executed actions:
{prev\_actions}
(END OF PREVIOUS ACTIONS)
Question: {question}
{scratchpad}

\end{quote}

\end{document}